%% file: main.tex
\documentclass[runningheads]{llncs}

\usepackage{sty}

\usepackage{abbrv}

\usepackage{graphicx}
\usepackage{booktabs}
\usepackage{textcomp} 

\usepackage[accsupp]{axessibility}

\usepackage{pifont}

\newcommand{\xmark}{\ding{55}}

\usepackage{hyperref}

\begin{document}

\title{Just Zoom In: Cross-View Geo-Localization via Autoregressive Zooming}

\titlerunning{Just Zoom In for Cross-View Geo-Localization}

\author{Yunus Talha Erzurumlu\inst{1} \and
Jiyong Kwag\inst{1} \and
Alper Yilmaz\inst{1}}

\authorrunning{Y.\,T.\ Erzurumlu et al.}

\institute{Photogrammetric Computer Vision Lab, The Ohio State University\\
\email{\{erzurumlu.1,kwag.3,yilmaz.15\}@osu.edu}
}

\maketitle

\begin{abstract}
Cross-view geo-localization (CVGL) is the task of estimating a camera’s location by matching a street-view image to geo-referenced overhead imagery, enabling GPS-denied localization and navigation. Existing methods almost universally pose CVGL as an image-retrieval problem in a contrastively trained embedding space. This formulation ties performance to large batches and hard negative mining, and it ignores both the geometric structure of maps and the coverage mismatch between street-view and overhead imagery. In particular, salient landmarks visible from the street-view can fall outside a fixed satellite crop, making retrieval targets ambiguous and limiting explicit spatial inference over the map.
We propose \textbf{Just Zoom In}, an \emph{alternative} formulation that performs CVGL via autoregressive zooming over a city-scale overhead map. Starting from a coarse satellite view, the model takes a short sequence of zoom-in decisions to select a \emph{terminal satellite cell} at a target resolution, without contrastive losses or hard negative mining. We further introduce a realistic benchmark with crowd-sourced street views and high-resolution satellite imagery, reflecting real capture conditions. On this benchmark, Just Zoom In achieves state-of-the-art performance, improving \textbf{Recall@1 $<$ 50m} by \textbf{5.5\%} and \textbf{Recall@1 $<$ 100m} by \textbf{9.6\%} over the strongest contrastive-retrieval baseline, demonstrating the effectiveness of sequential coarse-to-fine spatial reasoning for cross-view geo-localization.
\end{abstract}

\section{Introduction}
\label{sec:intro}
Visual localization estimates a camera's  location by matching a captured image to a known environment. Specifically, cross-view geo-localization (CVGL) tackles this problem by using geo-referenced satellite imagery to localize a street-view query image~\cite{Lin2014CVGL, Workman2015WideArea}. Satellite imagery offers dense, large-scale coverage of outdoor environments and is widely available for many regions, making CVGL attractive for GPS-denied localization and navigation. The core difficulty in CVGL is bridging the extreme viewpoint differences between street-view and satellite imagery.

\begin{figure}[t]
  \centering
  \includegraphics[width=1\linewidth]{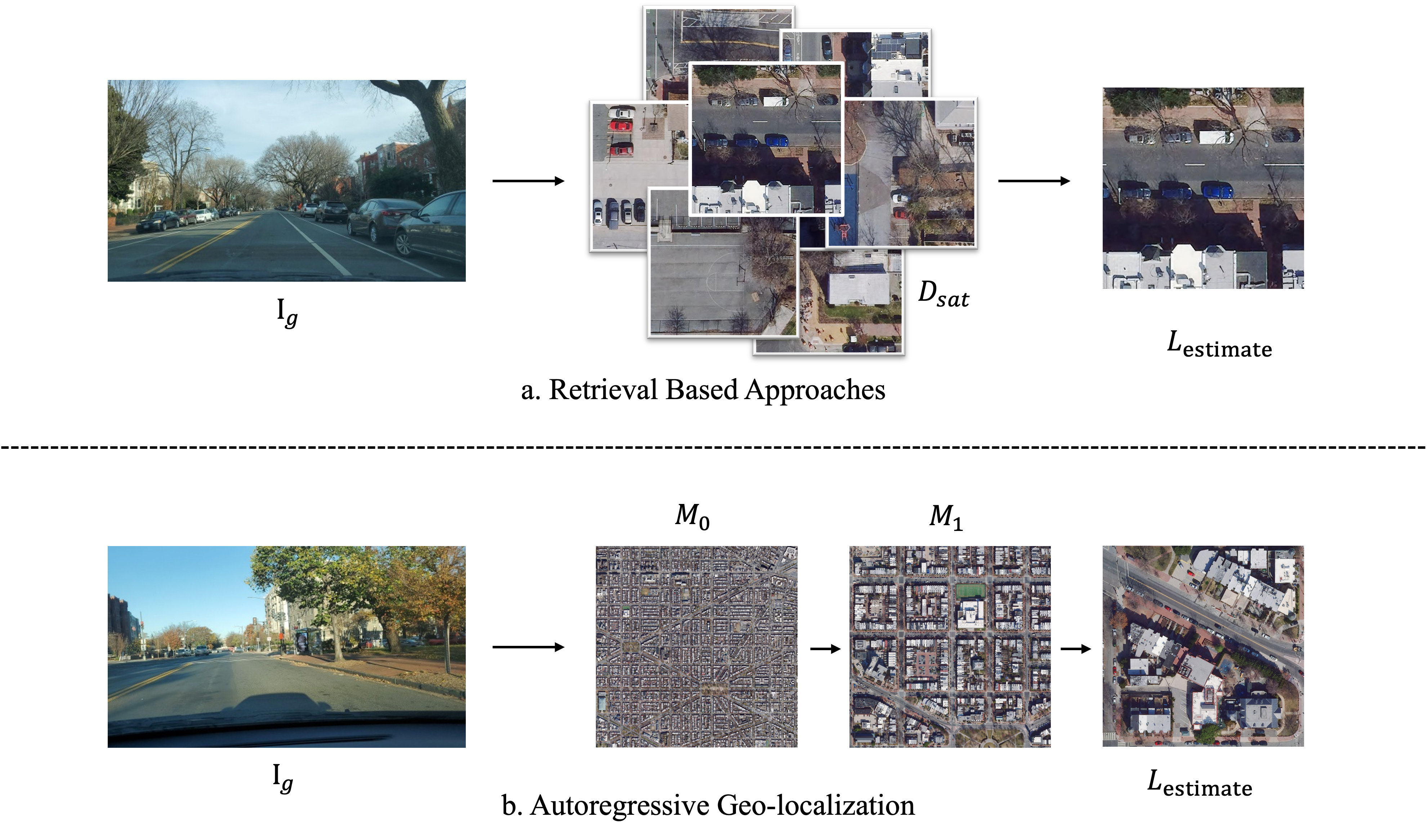}
  \caption{\textbf{Two paradigms for cross-view geo-localization.}
  (a) Retrieval-based approaches maintain a large GPS-tagged reference database and use a contrastively trained encoder to perform nearest-neighbor search at query time.
  (b) \emph{Autoregressive geo-localization} (ours) replaces global retrieval with sequential, coarse-to-fine zoom-in decisions over multi-scale satellite imagery, dramatically reducing dependence on exhaustive database search.}
  \label{fig:onecol}
\end{figure}

To handle these viewpoint differences, modern CVGL systems predominantly rely on contrastively trained retrieval models; given a street-view image, the system retrieves the most similar satellite tile from a large GPS-tagged reference database~\cite{Shi2019SAFA, Deuser2023Sample4Geo, Zhu2021VIGOR, Zhu2022TransGeo}. In practice, they learn a single embedding space in which street-view queries and satellite tiles are pulled together if they correspond to the same location and pushed apart otherwise, commonly relying on many informative negatives (via large effective batch sizes and/or hard negative mining). At inference time, localization is performed by nearest-neighbor search in the embedding space over a large satellite database shown in Figure~\ref{fig:onecol}.a.

Despite strong performance, this contrastive--retrieval formulation has three practical limitations for scalable, realistic CVGL. 
First, contrastive retrieval performance often benefits from many informative negatives, implemented via large effective batches and/or hard negative mining (HNM)~\cite{Deuser2023Sample4Geo, Zhu2022TransGeo}. Maintaining and refreshing hard negatives can add significant training compute and engineering complexity~\cite{Berton2022CosPlace}. 

Second, retrieval-based CVGL must store and search a large GPS-tagged reference database at inference time. Fine-grained localization over a city-scale region requires many densely sampled satellite tiles, so memory footprint grows roughly linearly with the number of tiles~\cite{Zhu2021VIGOR, Zhu2022TransGeo, Deuser2023Sample4Geo}, making it expensive to extend to larger areas.

\begin{figure}[t]
  \centering
  \includegraphics[width=1\linewidth]{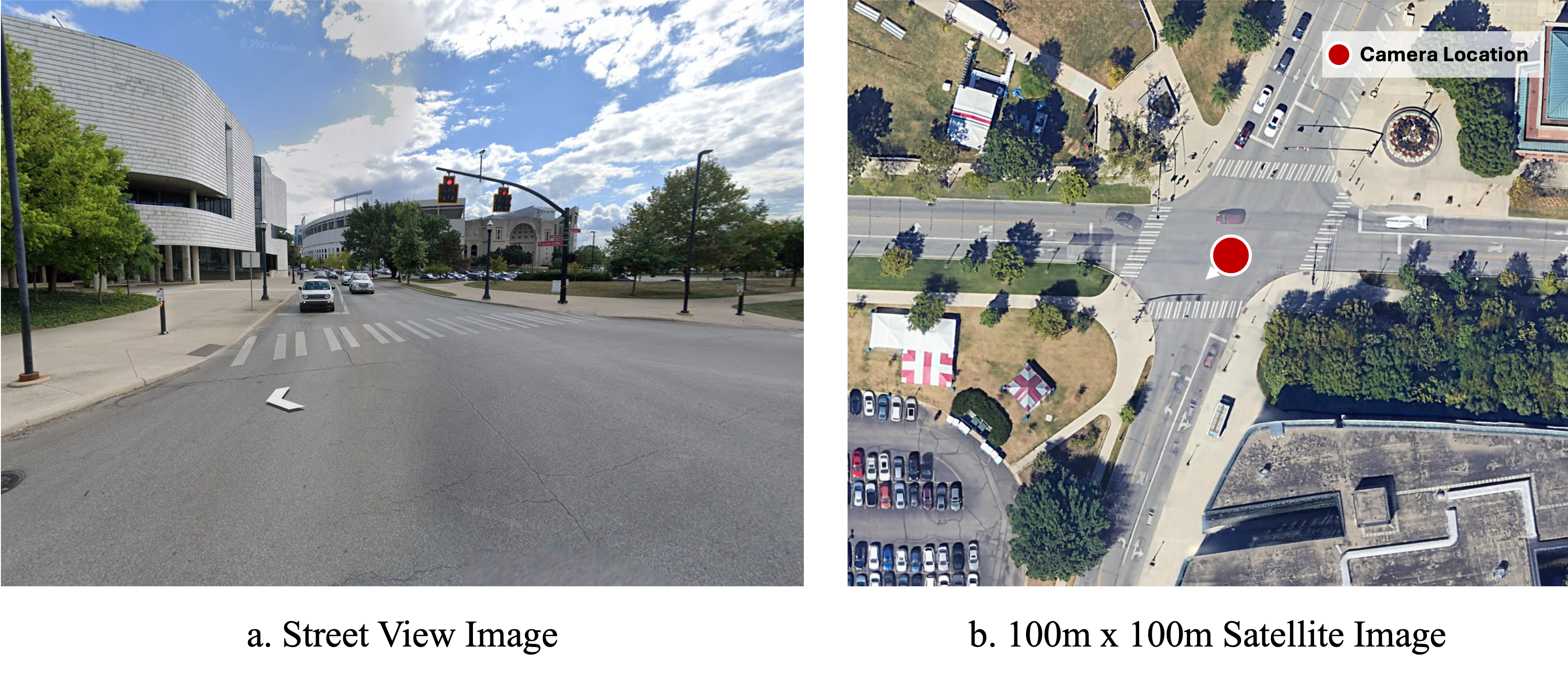}
  \caption{\textbf{Coverage mismatch.} In the street-view image (a), the  stadium and its distinctive gate are clearly visible. In the satellite crop (b) from the same region, this landmark falls outside of the patch, illustrating how small tiles can miss critical cues present in the street-view. Images sourced from Google Street View and Google Maps~\cite{google_maps_static_api, google_streetview_static_api}.}
  \label{fig:coverage}
\end{figure}

Third, the retrieval formulation treats the map as an unstructured collection of independent, fixed-size tiles, without explicit links between neighboring regions or coarser zoom levels. This tile-centric view ignores the underlying geographic hierarchy and can exacerbate a \emph{coverage mismatch} between modalities: due to perspective and tile framing/misalignment, discriminative context visible in the street-view image may lie outside a candidate satellite patch even when the patch is geographically close (Figure~\ref{fig:coverage}). As a result, existing methods are optimized to rank isolated patches rather than to reason over spatial context and multi-scale structure in the satellite imagery.

To address these limitations, we reformulate cross-view geo-localization as \emph{autoregressive zooming} rather than single-shot retrieval from a flat database. Our method, \emph{Just Zoom In} (Figure~\ref{fig:onecol}.b), localizes a ground query by producing a short sequence of zoom-in actions over a multi-scale satellite map, starting from a coarse view and refining the prediction until a terminal resolution. Motivated by 1D autoregressive prediction in NLP, we cast each zoom decision as a categorical token: at each of the $N$ zoom levels, the current satellite tile is subdivided into $K^2$ candidates, and the model predicts the next patch index conditioned on the street-view image and previous decisions. We instantiate this framework with an off-the-shelf visual encoder and a Transformer decoder~\cite{Vaswani2017Attention} trained with \emph{causal} masking and a supervised next-action (next-index) prediction objective, avoiding contrastive losses, explicit hard-negative mining, and large effective batch sizes. The resulting coarse-to-fine process explicitly follows the geographic hierarchy of the map: early steps leverage wide-area context, while later steps focus on local detail, helping mitigate coverage mismatch between street-level and satellite imagery. Since each step compares the query to only a few candidates, inference cost scales with the number of zoom steps rather than the total number of tiles in a city-scale database, reducing reliance on expensive nearest-neighbor search while achieving strong localization accuracy under realistic capture conditions.

This sequential formulation calls for a benchmark tailored to multi-scale reasoning.
To evaluate autoregressive zooming, we require satellite imagery with an explicit resolution hierarchy and street-views that resemble real capture conditions, such as limited field-of-view (FoV), unknown orientation, and diverse quality.
Existing datasets~\cite{Workman2015WideArea,Liu2019CVACT,Zhu2021VIGOR}, however, provide only single-scale satellite images paired with standardized 360\textdegree{} panoramas of known orientation and fixed intrinsics, which simplifies the task and departs from typical first-person or dash-cam imagery. Recent work further shows that performance can degrade substantially when moving from these idealized, orientation-aligned protocols to unknown-orientation and limited-FoV settings~\cite{Mi2024ConGeo}.
We therefore construct a new benchmark with multi-scale satellite images and randomly oriented, limited FoV street views captured under diverse conditions shown in Figure~\ref{fig:dataset_diversity}, so that both modalities match the demands of a realistic multi-scale zoom-in setting.

\paragraph{Scope and comparability.}
We target the \emph{retrieval stage} of city-/AOI-scale CVGL: given a ground query and a large overhead search region, the goal is to select the correct overhead \emph{cell} at a target resolution (Sec.~\ref{sec:datasets_metric}). Unlike cross-view \emph{metric pose estimation}, which assumes a coarse location prior (e.g., 20--50m) and then refines 3-DoF pose, we assume no such prior and output a coarse cell intended to \emph{provide} that prior for downstream refinement~\cite{Sarlin2023OrienterNet,Wu2024MapLocNet,Lentsch2023SliceMatch}. We further distinguish our goal from \emph{hierarchical} end-to-end pipelines that couple coarse retrieval with metric refinement in a single system: our method targets and strengthens the retrieval component that such pipelines depend on, and can be used as their front-end~\cite{Song2025UnifyGeo}.
We also depart from common retrieval-style CVGL protocols that rely on 360\textdegree{} panoramas and known/normalized orientation (e.g., North alignment), which can act as shortcuts; ConGeo shows performance can drop sharply, especially under limited-FoV, unknown-heading, and cross-area evaluation, highlighting that many models do not transfer to realistic user imagery~\cite{Mi2024ConGeo}. Our focus is the \emph{within-AOI (same-area)} setting where the localization region is supervised during training, enabling region-specific learning over a multi-scale overhead hierarchy. Generalizing to \emph{unknown areas} under realistic constraints remains substantially harder; recent active geo-localization work explores this direction, typically on smaller AOIs/coarser discretizations and with much heavier training machinery (billion-scale pretrained models and RL), whereas we show that supervised next-action zooming with a standard vision backbone is competitive in the supervised within-AOI regime~\cite{Sarkar2024GOMAA,Mi2025GeoExplorer}.

\paragraph{Contributions.} Our main contributions are:
\begin{itemize}
\item \textbf{Reformulating CVGL.} Motivated by the addressed limitations we reformulate cross-view geo-localization as \emph{ autoregressive zooming}.
\item \textbf{An autoregressive zoom-in model.} We propose \textbf{Just Zoom In}, an autoregressive CVGL transformer trained via supervised next-action prediction. Our design achieves computational efficiency while explicitly leveraging hierarchical visual cues from coarse to fine satellite scales.
\item \textbf{A realistic, multi-scale benchmark.} We introduce a cross-view dataset that pairs multi-scale satellite imagery with crowd-sourced street-view images exhibiting unknown orientation, limited and variable FoV, and diverse appearance conditions. Our approach achieves state-of-the-art performance on this benchmark, with \textbf{+5.50\%} and \textbf{+9.63\%} absolute gains at R@50m and R@100m over the \emph{state-of-the-art} contrastive-retrieval baseline.
\end{itemize}

\section{Related Works}
\label{sec:related}
\textbf{The Dominant Retrieval Paradigm.}
Cross-view geo-localization addresses visual localization by leveraging geo-referenced satellite imagery, which is valued for its extensive global coverage, currency, and accessibility. Early studies in this area formulated the problem using contrastive learning and image-retrieval objectives ~\cite{Lin2014CVGL, Workman2015WideArea}. This formulation remains the dominant approach in the literature to date ~\cite{Hu2018CVMNet, Cai2019Ground2Aerial,Shi2019SAFA, Regmi2019Bridging, Liu2019CVACT,Zhu2021VIGOR,Zhu2022TransGeo,Deuser2023Sample4Geo,Zhu2023SAIG,Shugaev2024ArcGeo}. Because satellite and street-level views differ dramatically, the task requires robust feature extraction ~\cite{Workman2015WideArea,Zhu2022TransGeo, Zhu2023SAIG}, principled aggregation ~\cite{Arandjelovic2015NetVLAD,Shi2019SAFA}, and well-designed sampling ~\cite{Cai2019Ground2Aerial,Deuser2023Sample4Geo}.

\begin{figure}[t]
  \centering
  \includegraphics[width=\textwidth]{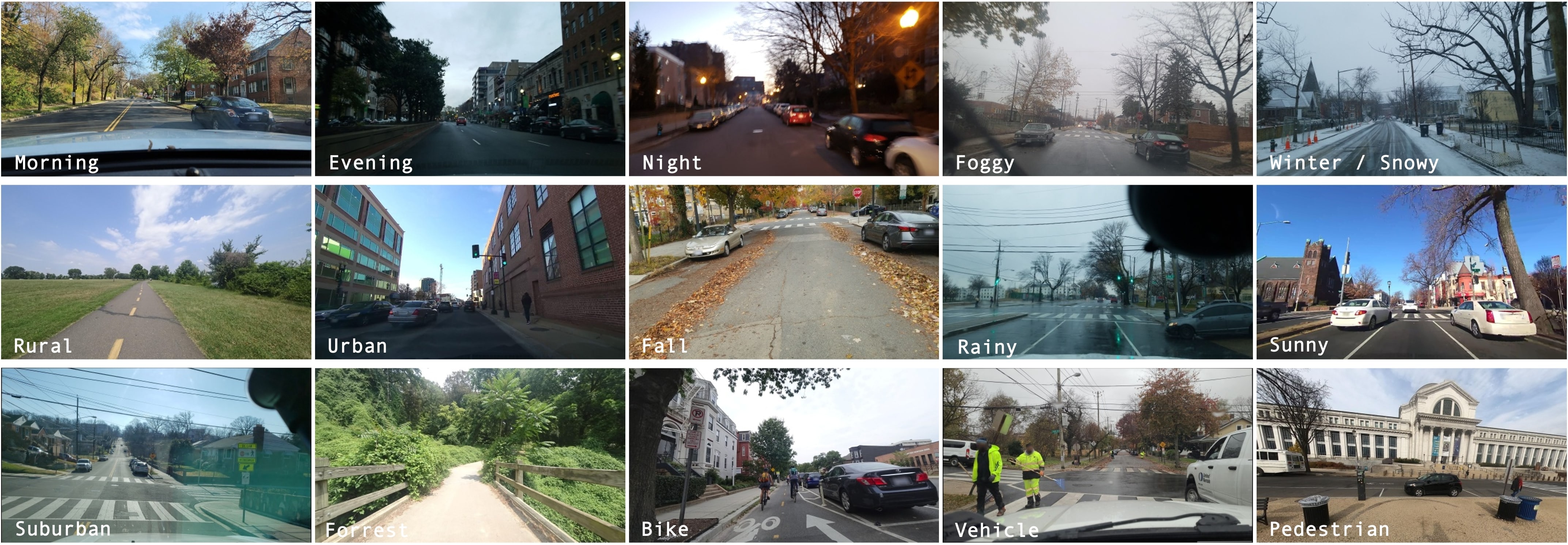}
  \caption{\textbf{Dataset examples.} Street-view samples from our cross-view image localization corpus illustrating the two defining characteristics: (i) \emph{limited  FoV} typical of first-person and dash-cam capture; and (ii) broad diversity across time of day, seasonal appearance, weather, scene type , and capture platforms . This variability-together with viewpoint changes, occlusions, and motion blur-widens the appearance gap to overhead imagery and makes cross-view matching more challenging.}
  \label{fig:dataset_diversity}
\end{figure}

On the feature-extraction front, Workman~\etal~\cite{Workman2015WideArea} showed that CNN-learned features~\cite{Krizhevsky2012AlexNet} outperform hand-crafted descriptors ~\cite{Lin2014CVGL}. Building on this, TransGeo~\cite{Zhu2022TransGeo} reported further gains by employing Vision Transformers ~\cite{Dosovitskiy2021ViT}, and Sample4Geo~\cite{Deuser2023Sample4Geo} adopted ConvNeXt~\cite{Liu2022ConvNeXt}. Regarding aggregators, CVMNet~\cite{Hu2018CVMNet} leverages NetVLAD ~\cite{Arandjelovic2015NetVLAD}; SAFA ~\cite{Shi2019SAFA} introduces a customized spatial feature aggregator; and SAIG ~\cite{Zhu2023SAIG} utilizes an MLP-Mixer ~\cite{Tolstikhin2021MLPMixer} architecture. The largest performance boosts have stemmed from sampling:  HNM proved transformative. In particular, Sample4Geo~\cite{Deuser2023Sample4Geo} demonstrated marked improvements in its sampling approach. However, HNM makes training both compute-heavy and memory-hungry, highlighting a central limitation of contrastive learning. In contrast, our formulation attains state-of-the-art performance without specialized sampling. Some studies attempted to close the view-point gap through polar transforms~\cite{Shi2019SAFA} or generative methods ~\cite{Toker2021ComingDown, Regmi2019Bridging}, but received little uptake owing to unrealistic assumptions, particularly the requirement that the camera position in the satellite imagery be known.
\input{tables/dataset_comp}
\textbf{Benchmark and Dataset Limitations.} Another crucial aspect of this task is the dataset and the realism of the benchmarks. Early datasets such as CVUSA~\cite{Workman2015WideArea} and CVACT~\cite{Liu2019CVACT} used one-to-one pairing for data generation, i.e., a single street-view image sampled at the center of each satellite tile. This renders the task both trivial and unrealistic: the reference database becomes sparsely populated over the test region, hindering the localization of novel photos. VIGOR~\cite{Zhu2021VIGOR} addressed this limitation by proposing many-to-one sampling, in which multiple street-view images are sampled from random locations within each satellite tile. This setting is more realistic since the exact location of the street view capture is unknown and provides better coverage. However, in VIGOR and other datasets, the street-view orientations are known, and the FoVs are 360$^\circ$, which simplifies the task and again departs from real-world conditions. Moreover, the dataset is single-sourced and highly standardized, which can yield models that generalize poorly to out-of-distribution data and make the benchmark easier. By contrast, while we also adopt many-to-one sampling, we use random orientations and limited, randomly selected FoVs, and we draw street-view images from diverse sources, times of day, seasons, weather conditions, and quality levels, resulting in a more challenging and realistic benchmark as illustrated in Figure~\ref{fig:dataset_diversity}. Recent work considers limited FoV and unknown orientation~\cite{Shugaev2024ArcGeo, Mi2024ConGeo}, but largely relies on panoramic training data with simulated FoV/heading, whereas our benchmark uses non-panoramic, randomly oriented, limited FoV images, hence the settings are not directly comparable.


\textbf{Cross-View Pose Estimation.}
A parallel line of work to CVGL is cross-view pose estimation, where a coarse location prior (typically 20–50m) is available, and the goal is to recover a 3-DoF camera pose with sub-meter accuracy~\cite{Sarlin2023OrienterNet,Wu2024MapLocNet,Lentsch2023SliceMatch}. Although recent methods target this setting~\cite{tong2025geodistill, Lee2025PIDLoc, Xia2025FG2}, its practicality hinges on a reliable CVGL component to furnish the coarse prior and robust cross-view matches-precisely what our approach aims to provide.

\section{Method}
\label{sec:method}
In this section we will discuss how we curated high quality multi-scale dataset, \ref{sec:dataset}, and the theoretical and architectural details of \emph{Just Zoom In} Model, \ref{sec:model}.
\subsection{Multi-Scale Satellite and Street-View Dataset}
\label{sec:dataset}
The development of our autoregressive geo-localization framework requires a bespoke dataset that satisfies two key criteria: (1) realistic variability in street-view imagery (\ref{sec:ground}), and (2) multi-scale, high-resolution satellite imagery (\ref{sec:sat}). Since no existing open-source cross-view geo-localization benchmark meets these requirements, we constructed a dedicated dataset by combining crowd-sourced street-view imagery from Mapillary~\cite{mapillary} with government-sourced high-resolution aerial orthophotography~\cite{opendatadc}.

\subsubsection{Street-view processing.}
\label{sec:ground}
Mapillary~\cite{mapillary} is selected as the source of street-view query images due to its inherent representation of in-the-wild visual variability. Mapillary is a large-scale crowd-sourced public dataset licensed under CC BY-SA, containing more than 2 billion images. The platform provides images that span a range of fields of view captured by heterogeneous devices under diverse environmental conditions, including varying weather, seasons, time of day, and camera models shown in Figure~\ref{fig:dataset_diversity}. This diversity introduces realistic appearance changes that are often absent in curated or proprietary datasets, making it a suitable testbed for robust cross-view localization. We collect data from the Washington D.C. area, covering both dense urban centers and surrounding rural regions, resulting in approximately 300k street-view images. More details are included in the Appendix.

The crowd-sourced nature of Mapillary introduces noise such as low-resolution, outdated, or duplicate imagery. To convert raw sequences into a clean and geographically reliable dataset, we adopt a multi-stage filtering pipeline inspired by Map It Anywhere~\cite{ho2024map}. First, we discard images whose initial GPS coordinates significantly deviate from their Structure-from-Motion (SfM) refined poses, removing samples with inaccurate ground-truth locations. Second, to promote spatial diversity, we enforce a minimum spatial separation constraint by retaining only key frames, ensuring that no two consecutive images are within 4.0 meters of each other. Third, we filter images based on temporal and device metadata to exclude outdated or unreliable camera sources, increasing the diversity of environmental and seasonal conditions represented. Finally, we perform quality screening following the Map It Anywhere pipeline, discarding images with motion blur, lens distortion, heavy occlusion, or reflections. Non-panoramic images are standardized to a resolution of $512 \times 384$.

\subsubsection{Satellite-view processing.}
\label{sec:sat}
For aerial reference data, we use publicly available government-sourced orthophotos~\cite{opendatadc}. These datasets provide high-resolution imagery distributed under the permissive CC BY 4.0 license, ensuring reproducibility and scalability. The aerial data for the Washington D.C. region is obtained from the official open government data portal similar to Fervers~\etal~\cite{fervers2024statewide}, covering a $10\,\mathrm{km} \times 10\,\mathrm{km}$ area that corresponds to the spatial distribution of street-view images from Mapillary.

To support coarse-to-fine zooming, we organize the satellite imagery into a hierarchy of tiles. At each zoom level we subdivide the current tile into a fixed grid of $K \times K = 4 \times 4$ patches, and the model chooses one patch to zoom into. In all experiments we use $N = 4$ zoom steps (i.e., four actions), which correspond to nominal tile footprints of
$10\,\mathrm{km} \times 10\,\mathrm{km}$,
$2.5\,\mathrm{km} \times 2.5\,\mathrm{km}$,
$625\,\mathrm{m} \times 625\,\mathrm{m}$, and
$156.25\,\mathrm{m} \times 156.25\,\mathrm{m}$.
After the final zoom decision, we take the center $39.06\,\mathrm{m} \times 39.06\,\mathrm{m}$ region inside the selected $156.25\,\mathrm{m}$ tile as the output cell used for evaluation. This finest scale is only used to define the prediction footprint and to compute geodesic distances; it is not supplied to the model as an input image.

We provide a comparison with our new dataset with existing ones at Table~\ref{tab:data_comp}.

\begin{figure}[t]
  \centering
  \includegraphics[width=\textwidth]{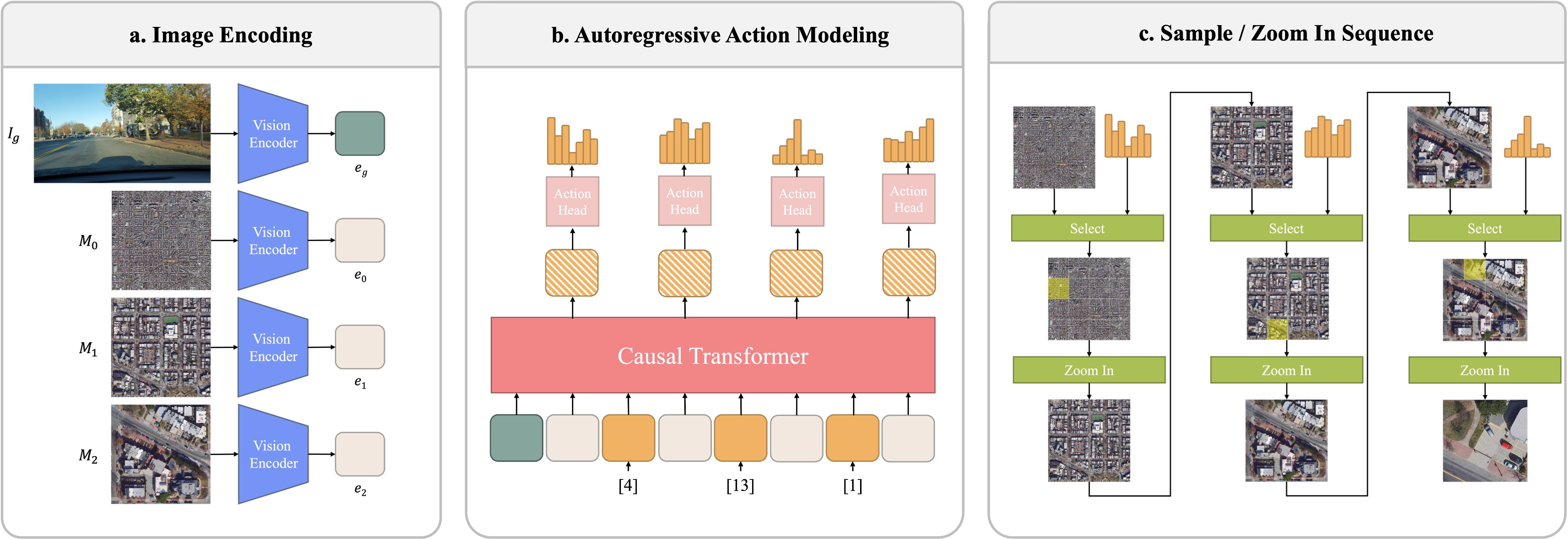}
  \caption{Overview of \emph{Just Zoom In}. (a) \textbf{Image encoding.} A shared, off-the-shelf vision encoder maps the street-view image \(I_g\) and the multi-scale satellite maps \(\{M_t\}\) to global representation tokens \(e_g\) and \(e_t\), single token per image. (b) \textbf{Autoregressive action modeling.} The image tokens, interleaved with previously chosen action tokens, are fed to a causal transformer that predicts a distribution over the next zoom-in action \(a_t\) at each step. (c) \textbf{Sample / zoom-in sequence.} Beginning from \(M_0\), the model selects the most probable patch, zooms to obtain \(M_{t+1}\), and iterates until a terminal patch \(M_N\); the center of \(M_N\) is taken as the location estimate.}
  \label{fig:overview_model}
\end{figure}

\subsection{Just Zoom In}
\label{sec:model}
The autoregressive zooming process is modeled as a sequential decision-making problem. Given a street-view image $I_g$, the goal is to predict the correct satellite patch $M_N$ at each scale by learning a policy $\pi$ that produces a sequence of $N$ zoom-in actions. Our autoregressive geo-localization framework consists of two main components. First, as shown in \ref{fig:overview_model}.a the shared-weight vision encoder bridges the domain gap between street-view and satellite images, projecting them into a unified feature space. Second, the autoregressive decoder~\cite{Vaswani2017Attention} receives the street-view feature along with multi-scale satellite features and, under a causal masking strategy, predicts the next zoom action step-by-step, as illustrated in \ref{fig:overview_model}. We will discuss these components in more detail in the following sections.

\subsubsection{Vision Encoder.}
\label{sec:vision_enc}
The first stage of our model is responsible for bridging the significant domain gap between street-view and satellite views. To tackle this, we employ the DINOv2-Base model~\cite{Oquab2023dinov2} as our vision encoder backbone. DINOv2 is a powerful foundation model pre-trained using self-supervised learning for general-purpose visual representations. We use a shared-weight DINOv2 encoder to process both the street-view image and all multi-scale satellite images. This weight sharing encoder projects both modalities into a common embedding space, effectively reducing the domain gap caused by drastic viewpoint differences.

For each street-view or satellite image, we extract only the \texttt{[CLS]} token embedding as the output representation. The \texttt{[CLS]} token is designed to aggregate global information from the entire image into a compact vector. Street-view and multi-scale satellite images are encoded and concatenated to form $[I_g, M_0, M_1, \dots, M_N]$ for autoregressive model input. To preserve the strong pre-trained visual knowledge while allowing adaptation to our specific task domain, we employ a partial fine-tuning strategy. The encoder is initialized with pre-trained DINOv2 weights, and only the last four transformer blocks are unfrozen and fine-tuned during training, a parameter-efficient transfer approach shown to be effective for localization tasks ~\cite{izquierdo2024optimal}.
\input{tables/main_comp}

\subsubsection{Autoregressive Geo-Localization Decoder.}
\label{sec:arg}
The model predicts a discrete zoom-action sequence $Y = \{y_0,\dots,y_{N-1}\}$, where each $y_t \in \{0,\dots,K^2-1\}$ selects one of the $K^2$ child tiles at zoom step $t$.
We train with teacher forcing, maximizing the likelihood of the ground-truth action at each step (equivalently, minimizing cross-entropy over the $K^2$ action classes):
\begin{equation}
\mathcal{L} = -\sum_{t=0}^{N-1} \log P\!\left(y_t^{*}\mid I_g, M_0, y_0^{*}, \dots, M_t\right),
\end{equation}
where $(\cdot)^{*}$ denotes ground-truth actions/tokens. The action tokens are interleaved with multi-scale satellite features to form the input sequence $[I_g, M_0,\allowbreak y_0^{*},\allowbreak M_1,\allowbreak y_1^{*},\allowbreak \dots,\allowbreak M_{N-1}]$. This interleaving design ensures that each zoom decision is conditioned on the correct patch information, providing stronger guidance for subsequent predictions. The transformer decoder with 1D causal masking ensures that each token attends only to itself and the past elements, preventing information leakage from future zoom steps.

During inference, using the chain rule of probability, the joint objective factorizes into step-wise conditional probabilities:
\begin{equation}
P(Y \mid I_g, M_0) = \prod_{t=0}^{N-1} P(y_t \mid I_g, M_0, y_0, \dots, M_t).
\label{eq:autoregressive_policy}
\end{equation}
The autoregressive nature requires that each prediction be conditioned on previous history decisions. A key advantage of this setup is that the search space is progressively reduced at every zoom step. Figure \ref{fig:overview_model}.b shows an overview of the sequential zoom-in mechanism.

\section{Experiments}
\label{sec:experiments}

Training and evaluation are conducted on our curated dataset described in Section~\ref{sec:dataset}.

\subsection{Metrics}
\label{sec:datasets_metric}
Unlike standard image retrieval tasks that rely on rank-based metrics, our iterative zooming approach produces a single final location prediction. Furthermore, simple correct/incorrect classification accuracy is insufficient for geospatial tasks, since predicting a neighboring patch is considerably more accurate than predicting a distant one. Therefore, we evaluated performance using distance-based Recall@1 (R@1 $< d$). Let $d(\hat{c}_i, c_i)$ denote the geodesic distance in meters between the predicted center $\hat{c}_i$ and the ground-truth center $c_i$ for sample $i$. The Recall@1 with a success radius $\tau \in \{40\text{m},\, 50\text{m},\, 100\text{m}\}$ is defined as:
\begin{equation}
\text{R@}\tau\text{m} = \frac{1}{S} \sum_{i=1}^{S} \mathbf{1}\!\left(d(\hat{c}_i, c_i) \le \tau\right) \times 100,
\end{equation}
where
\begin{equation}
\mathbf{1}(x) =
\begin{cases}
1, & \text{if } x \text{ is true}, \\
0, & \text{otherwise.}
\end{cases}
\end{equation}

\subsection{Implementation Details}
\label{sec:implementation}
The autoregressive transformer blocks are built using modern components for training stability and performance. The model operates with a hidden dimension of $768$, using $6$ layers and $8$ attention heads. We adopt a pre-normalization structure with RMSNorm~\cite{Biao2019RMSNorm} inspired by the LLaMA-3 family ~\cite{grattafiori2024llama}, where normalization is applied before the attention and feed-forward layers. Rotary positional embeddings (RoPE) ~\cite{su2023roformerenhancedtransformerrotary} are applied in 1D to enforce the sequential structure of the prediction process. We use AdamW~\cite{loshchilov2018decoupled} optimizer with a learning rate of $3\times10^{-4}$ and apply gradient clipping with a norm threshold of 1.0 to prevent exploding gradients. Training is conducted on two NVIDIA A6000 GPUs with a batch size of 128 for 30 epochs.

\begin{figure}[t]
  \centering
  \includegraphics[width=\linewidth]{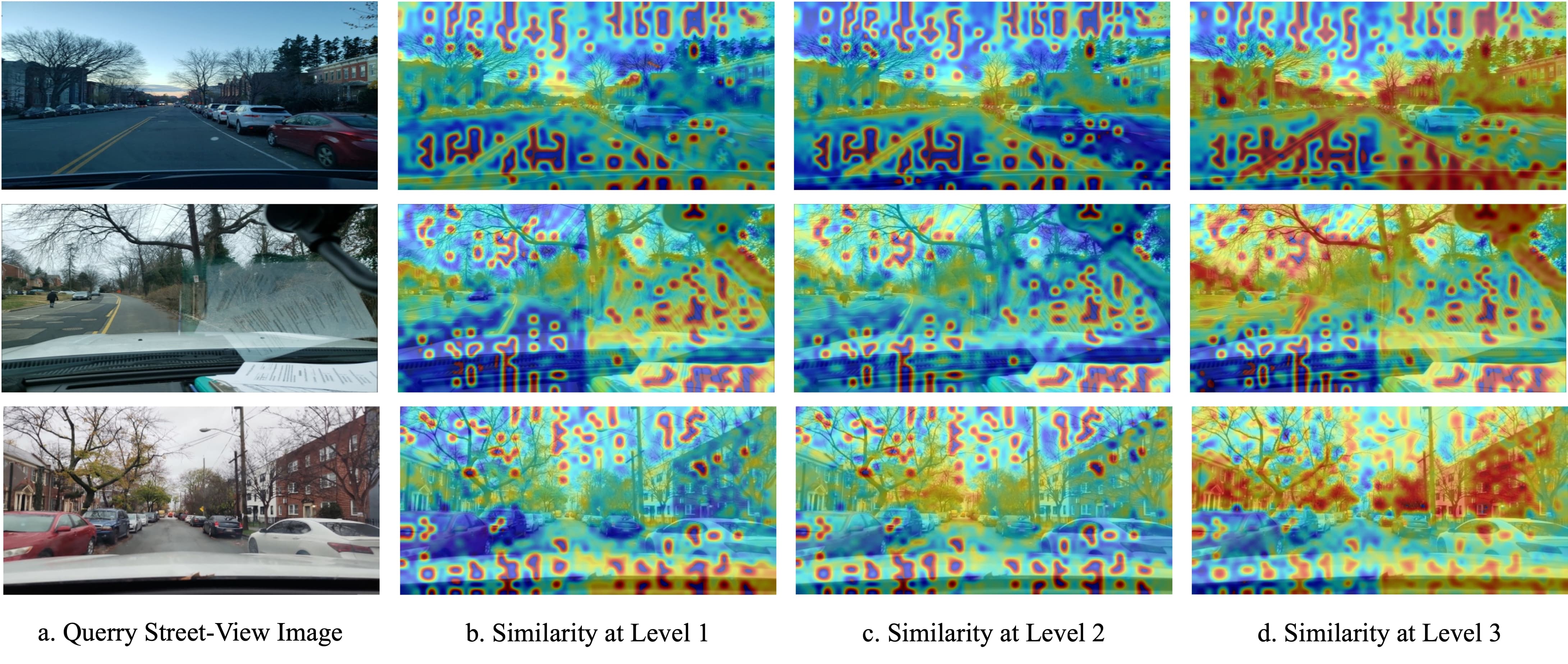}
  \caption{\textbf{Focus at different zoom levels.} We visualize similarity maps between street-view patch tokens and the satellite \texttt{[CLS]} embedding (blue = low similarity, red = high similarity) across three zoom levels. At the coarsest level, high similarity concentrates on far-away regions, whereas at finer zoom levels it shifts toward nearby objects and road/building details, indicating that the model learns to exploit different scale-specific visual cues.}
  \label{fig:sim_map}
\end{figure}

\subsection{Comparison with Contrastive Baselines}
\label{sec:result}
\begin{figure}[t]
  \centering
  \includegraphics[width=\textwidth]{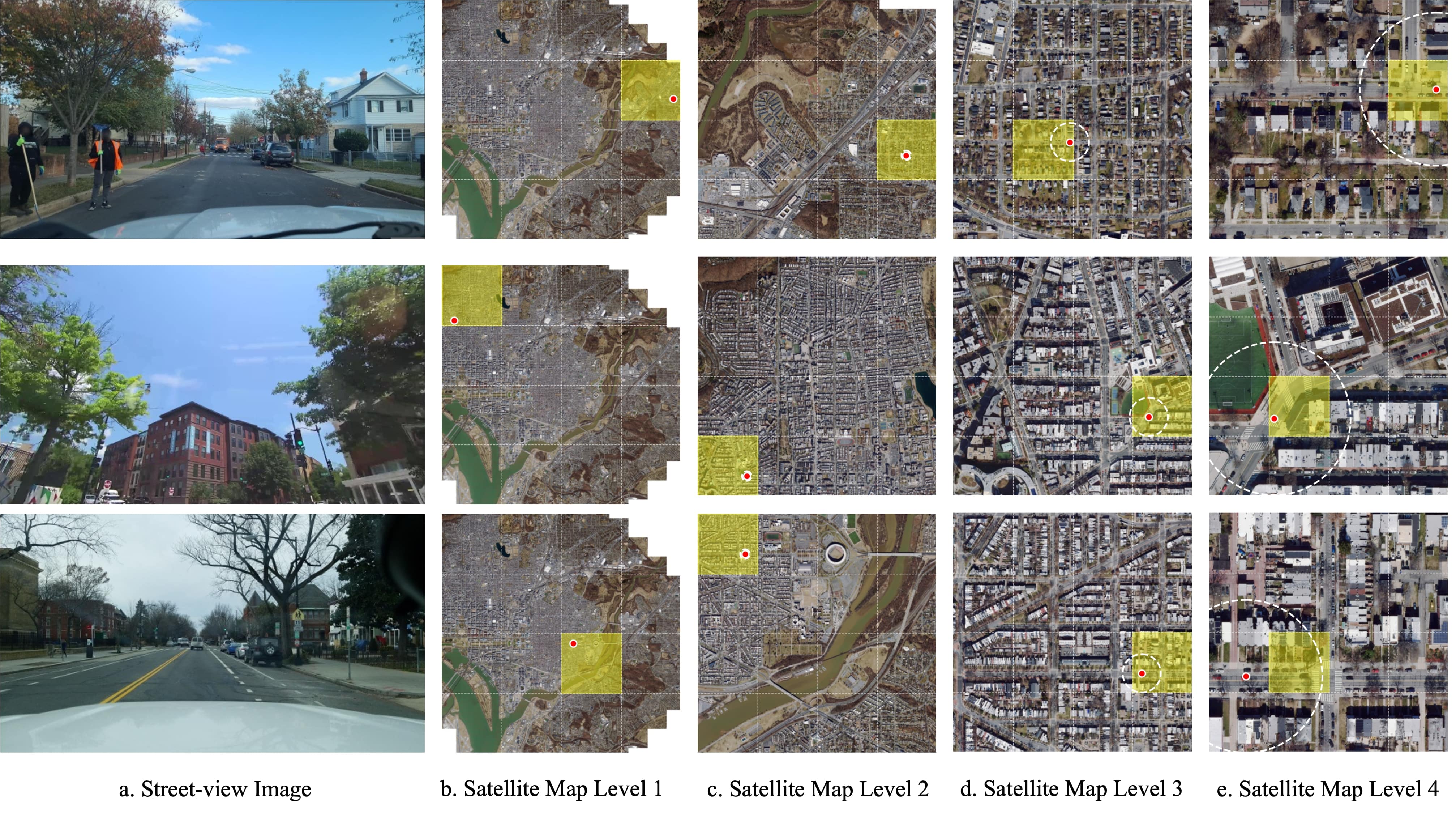}
  \caption{\textbf{Results.} \label{fig:qualitative_results} Qualitative visualization of our model's auto-regressive localization. The first column shows the input street-view query. The subsequent columns (Lvl 1-4) display the satellite map at each sequential zoom step. On each map, the \textbf{model's predicted patch} is highlighted by the grid overlay, the \textbf{ground truth (GT) location} is marked with a red dot, and the \textbf{50m success radius} is shown as a white dotted circle.}
  \label{fig:results}
\end{figure}

Table~\ref{tab:main_comp} presents a comparison between our autoregressive geo-localization approach and standard contrastive learning-based baselines with $10\,\mathrm{km} \times 10\,\mathrm{km}$. All baselines are reproduced using their official implementations, with additional training details provided in the Appendix. Our autoregressive framework consistently outperforms contrastive retrieval methods across all distance thresholds. Notably, our method achieves a +5.5\% absolute improvement in R@50m compared to the \emph{state-of-the-art} Sample4Geo~\cite{Deuser2023Sample4Geo} baseline with half the training time without needing to HNM. This performance advantage supports the core premise that an iterative, coarse-to-fine search strategy is more effective than single-shot retrieval for fine-grained localization. By progressively narrowing the search space, \textit{Just Zoom In} is able to leverage contextual cues at the appropriate spatial scale, avoiding the ``needle-in-a-haystack'' failure mode that often arises when retrieval systems must search large geographic regions directly. The substantial margin at R@100m (nearly +10\% over Sample4Geo~\cite{Deuser2023Sample4Geo}) further indicates that autoregressive geo-localization is highly robust, rarely producing catastrophic location errors that place the prediction in entirely incorrect neighborhoods. Moreover, as we can see from Figure \ref{fig:sim_map}, model learns to utilize different visual cues from the street-view image at different zoom levels, as expected, effectively dealing with the \emph{coverage mismatch} problem present in the contrastive-retrieval methods. Figure~\ref{fig:results} visualizes some of the results and zoom-in sequences.

\input{tables/enc_ablation_comp}
\input{tables/cls_comp}

\subsection{Ablation Study}
\label{sec:ablation}
To evaluate the contribution of each component, we conducted ablation studies on our dataset. For efficiency, all ablation experiments are performed using a $4 \times 4$ satellite patch layout with $N = 3$ zoom steps under a $2\,\mathrm{km} \times 2\,\mathrm{km}$ search setup.

\textbf{Vision encoder choices.}
The choice of a vision encoder is critical for bridging the significant domain gap between street-view and satellite imagery. To assess the inherent capability of different state-of-the-art vision foundation models for this task, we evaluate them as frozen feature extractors, training only the subsequent autoregressive policy module (Table~\ref{tab:enc_ablation}). Self-supervised models such as DINOv2~\cite{Oquab2023dinov2} and DINOv3~\cite{simeoni2025dinov3} significantly outperform the language-aligned SigLIP2~\cite{tschannen2025siglip2}. Although SigLIP2 retains some transferable understanding, its performance is insufficient for the fine-grained spatial discrimination required for accurate geo-localization. Based on these results, we select DINOv2 as the default backbone for all subsequent experiments.

\textbf{Vision encoder fine-tuning.}
Although foundation models such as DINOv2 provide strong general-purpose visual features, they are not natively aligned to drastic viewpoint changes between street-view and satellite imagery. To assess the need for task-specific adaptation, we experimented with different numbers of frozen layers during training. This allows the highest-level semantic features to adapt to the domain shift between street-view and satellite representations. As shown in Table~\ref{tab:enc_ablation}, the strictly off-the-shelf features are insufficient for high-precision localization. Increasing the number of trainable layers beyond the final six leads to diminishing returns. This observation aligns with the intuition that earlier layers of foundation models encode general geometric and texture-level features, whereas later layers capture higher-level semantic information that benefits from adaptation to cross-view correspondence~\cite{izquierdo2024optimal}.

\textbf{Global vs. local patch features.}
Our default model uses only the \texttt{[CLS]} token from the vision encoder, which serves as a global descriptor for the entire satellite patch. We also investigated whether incorporating local patch tokens from DINOv2 can improve fine-grained matching. To test this, we compare the single-\texttt{[CLS]} baseline against a variant that selectively includes the $S$ patch tokens most similar to the street-view \texttt{[CLS]} token. As shown in Table~\ref{tab:local_comp}, incorporating local patch tokens, even when filtered based on semantic similarity, leads to a performance decline. This result suggests that the autoregressive zooming policy benefits from a high-level global representation when making sequential decisions, and that introducing fine-grained local details may introduce noise or over-fit to appearance variations rather than stable structural cues.

\section{Conclusion}
We introduced Just Zoom In, an autoregressive framework for cross-view geo-localization that replaces single-shot contrastive retrieval with a sequential, coarse-to-fine localization process. By jointly encoding street-view and multi-scale satellite imagery into a shared representation and predicting zoom actions with a causal transformer, our method progressively narrows the search region while preserving global spatial context. This design enables efficient inference that scales logarithmically with geographic area, avoiding the computational and representation bottlenecks of contrastive matching. We further curated a new multi-scale dataset composed of government-sourced orthophotography and crowd-sourced street-view imagery, providing a realistic and reproducible benchmark for cross-view localization. Extensive experiments and ablations demonstrate that our autoregressive formulation yields consistently higher localization accuracy and robustness compared to state-of-the-art contrastive baselines, particularly in challenging real-world conditions. We hope this work motivates a shift from retrieval-based matching toward sequential spatial reasoning in cross-view localization.


\bibliographystyle{splncs04}
\bibliography{main}

\end{document}

%% file: tables/dataset_comp.tex
\begin{table}[t]
\centering
\caption{\textbf{Dataset comparisons for cross-view geo-localization.} We compare our proposed multi-scale benchmark against prior datasets under key capture and supervision attributes. \textit{SV GPS}: \emph{Aligned} = center-aligned one-to-one pairing; \emph{Arbitrary} = query location not constrained to the tile center. \textit{Gnd FoV}: \emph{360\textdegree{} pano} = panorama queries; \emph{Crops} = crops from panoramas; \emph{Perspective} = native non-panoramic images. \textit{Heading}: \emph{Fixed} = known/normalized orientation; \emph{Arbitrary} = unknown or random orientation.}
\label{tab:data_comp}

\resizebox{\linewidth}{!}{%
\begin{tabular}{l c c c c c c c c c}
\toprule
\textbf{Dataset} & \textbf{\# Photos} & \textbf{Multi-Scale} &
\multicolumn{3}{c}{\textbf{Ground / Pairing Protocol}} &
\multicolumn{4}{c}{\textbf{Domain Types}} \\
\cmidrule(lr){4-6} \cmidrule(lr){7-10}
 & & & \textbf{SV GPS} & \textbf{Gnd FoV} & \textbf{Heading} &
\textbf{Urban} & \textbf{Suburban} & \textbf{Rural} & \textbf{Offroad} \\
\midrule
CVUSA~\cite{Workman2015WideArea}        & 44K   & \xmark &
Aligned & 360\textdegree{} pano & Fixed &
\checkmark & \checkmark & \checkmark & \checkmark \\
CVACT~\cite{Liu2019CVACT}      & 128K  & \xmark &
Aligned & 360\textdegree{} pano & Fixed &
\checkmark & \checkmark & \xmark      & \xmark \\
Vo \& Hayes~\cite{VoHays2016Localizing} & 450K  & \xmark &
Aligned & Crops    & Arbitrary &
\checkmark & \checkmark & \xmark      & \xmark \\
VIGOR~\cite{Zhu2021VIGOR}              & 105K  & \xmark &
Arbitrary & 360\textdegree{} pano & Fixed &
\checkmark & \xmark      & \xmark      & \xmark \\
\midrule
Ours                                   & 300K  & \checkmark &
Arbitrary & Perspective & Arbitrary &
\checkmark & \checkmark & \checkmark & \checkmark \\
\bottomrule
\end{tabular}%
}
\end{table}

%% file: tables/main_comp.tex
\begin{table}[t]
\centering
\caption{\textbf{Cross-view geo-localization results on our proposed dataset.} We reproduce all contrastive-based baselines using their official implementations. Our coarse-to-fine autoregressive approach outperforms retrieval-based methods across all evaluation thresholds. \textit{R@$\tau$m} denotes Recall@1 within $\tau$ meters. \textit{Inf.\ Mem.\ (GB)} reports the peak memory required to store and search the reference database at inference time, \textit{Train.\ Time (GPU hours)} is the approximate end-to-end training time on one NVIDIA A6000 GPU, and \textit{HNM} indicates whether the method relies on hard negative mining during training.}
\label{tab:main_comp}

\resizebox{\linewidth}{!}{%
\begin{tabular}{l c c c c c c}
  \toprule
  \textbf{Method} & \textbf{R@40m} $\uparrow$ & \textbf{R@50m} $\uparrow$ & \textbf{R@100m} $\uparrow$ & \textbf{Inf. Mem. (GB)} $\downarrow$ & \textbf{Train. Time} $\downarrow$ & \textbf{HNM}\\
  \midrule
  SAIG-D~\cite{Zhu2023SAIG}           & 39.36 & 47.52 & 64.17    & 18 &  $>90$  & \checkmark\\
  TransGeo~\cite{Zhu2022TransGeo}     & 45.97 & 54.55 & 67.61    & 11 & $>90$ & \checkmark\\
  Sample4Geo~\cite{Deuser2023Sample4Geo} & 52.85 & 60.81 & 71.30 & 16 & $>90$ & \checkmark\\
  \midrule
  Ours (BS. = 32)                               & 51.08 & 61.49 & 77.43 & \textbf{9} & 61 &\xmark\\
  Ours (BS. = 64)                               & 54.10 & 64.74 & 79.74 & \textbf{9} & 58 &\xmark\\
  Ours (BS. = 128)                               & \textbf{55.74} & \textbf{66.31} & \textbf{80.93} & \textbf{9} & \textbf{52} &\xmark\\
  \bottomrule
\end{tabular}
}
\end{table}

%% file: tables/enc_ablation_comp.tex
\begin{table}[t]
\centering
\footnotesize
\setlength{\tabcolsep}{3pt}
\renewcommand{\arraystretch}{1.05}
\caption{\textbf{Vision encoder ablations.} \textbf{(a) Comparison of different vision backbones.} We evaluate different  vision encoder backbones for our autoregressive geo-localization task. Overall, DINOv2 achieves the strongest performance, outperforming more recent architectures. \textbf{(b) Impact of partial backbone fine-tuning.} We evaluate different degrees of layer freezing in the vision encoder for our autoregressive geo-localization task. Overall, unfreezing the last 4 layers yields the best performance, outperforming both fully frozen and fully fine-tuned configurations.}
\label{tab:enc_ablation}

\begin{minipage}[t]{0.49\linewidth}
\centering
\textbf{(a) Vision backbones.}\\%
\resizebox{\linewidth}{!}{%
\begin{tabular}{@{}lccc@{}}
\toprule
\textbf{Method} & \textbf{R@40m} $\uparrow$ & \textbf{R@50m} $\uparrow$ & \textbf{R@100m} $\uparrow$ \\
\midrule
SigLIP2-B~\cite{tschannen2025siglip2} & 28.42 & 33.23 & 45.43\\
DINOv3-B~\cite{simeoni2025dinov3}     & 63.88 & 69.44 & 79.74\\
\midrule
DINOv2-B~\cite{Oquab2023dinov2}       & \textbf{64.13} & \textbf{69.65} & \textbf{79.95}\\
\bottomrule
\end{tabular}%
}
\end{minipage}\hfill
\begin{minipage}[t]{0.49\linewidth}
\centering
\textbf{(b) Partial backbone fine-tuning.}\\
\resizebox{\linewidth}{!}{%
\begin{tabular}{@{}lccc@{}}
\toprule
\textbf{Method} & \textbf{R@40m} $\uparrow$ & \textbf{R@50m} $\uparrow$ & \textbf{R@100m} $\uparrow$ \\
\midrule
Fully Frozen  & 58.79 & 64.66 & 76.51\\
Last 4 Layers & \textbf{66.98} & \textbf{72.25} & \textbf{83.08}\\
Last 6 Layers & 65.69 & 70.64 & 82.56\\
All Layers    & 59.45 & 64.31 & 74.78\\
\bottomrule
\end{tabular}%
}
\end{minipage}
\end{table}

%% file: tables/cls_comp.tex
\begin{table}
\caption{\textbf{Comparison of different satellite image representations.} We compare global \texttt{[CLS]} token features with variants that incorporate local patch tokens from the vision encoder. Our results show that relying on the global \texttt{[CLS]} representation yields the highest localization accuracy.}
\label{tab:local_comp}
\centering
\begin{tabular}{l c c c}
    \toprule
    \textbf{Method} & \textbf{R@40m} $\uparrow$ & \textbf{R@50m} $\uparrow$ & \textbf{R@100m} $\uparrow$ \\
    \midrule
    \texttt{[CLS]} + 256 & 63.38 & 68.25 & 79.83\\
    \texttt{[CLS]} + 64  & 64.60 & 70.10 & 80.70\\
    \texttt{[CLS]} + 32  & 65.80 & 72.10 & 83.02\\
    \texttt{[CLS]} + 16  & 65.31 & 71.6 & 82.60\\
    \texttt{[CLS]} + 8   & 65.31 & 70.76 & 82.12\\
    \midrule
    \texttt{[CLS]} Only  & \textbf{66.98} & \textbf{72.25} & \textbf{83.08}\\
    \bottomrule
\end{tabular}
\end{table}